\documentclass[10pt,twocolumn,letterpaper]{article}

\usepackage{iccv}
\usepackage{times}
\usepackage{epsfig}
\usepackage{graphicx}
\usepackage{amsmath}
\usepackage{amssymb}
\usepackage[table]{xcolor}
\usepackage[title]{appendix}
\usepackage{siunitx}
\usepackage{arydshln}
\usepackage{booktabs}  
\usepackage{subfigure}
\usepackage[accsupp]{axessibility}  

\definecolor{lavender}{rgb}{0.9, 0.9, 0.98}
\definecolor{Gray}{gray}{0.95}

\newcommand{\real}{\mathbb{R}}

\newcommand{\loss}{\mathcal{L}}

\newcommand{\reg}{\mathcal{R}}
\newcommand{\w}{\mathcal{W}}
\newcommand{\wsmooth}{\tilde{w}}
\newcommand{\wplus}{\mathcal{W}+}

\newcommand{\mals}{\text{MaLS}}
\newcommand{\samf}{\text{SaMF}}

\newcommand{\decoder}{\mathbf{G}}
\newcommand{\refencoder}{\mathbf{E}_{ref}}
\newcommand{\audencoder}{\mathbf{E}_{aud}}
\newcommand{\faceencoder}{\mathbf{E}_{face}}

\usepackage[breaklinks=true,colorlinks,bookmarks=false]{hyperref}
\definecolor{Blue}{rgb}{0.152, 0.294, 0.925}
\hypersetup{citecolor=Blue}
\iccvfinalcopy 

\def\iccvPaperID{10448} 

\ificcvfinal\fi
\begin{document}
\title{StyleLipSync: Style-based Personalized Lip-sync Video Generation}
\author{Taekyung Ki$^{1\star}$ \quad Dongchan Min$^{2\star}$ \\
$^{1}$AITRICS, ~$^{2}$Graduate School of AI, KAIST\\
{\tt\small tkki@aitrics.com, alsehdcks95@kaist.ac.kr}\\
{\small \url{https://stylelipsync.github.io}}}

\maketitle

\ificcvfinal\thispagestyle{empty}\fi
\def\thefootnote{$\star$}\footnotetext{Equal contribution.}\def\thefootnote{\arabic{footnote}}

\begin{abstract}
In this paper, we present \textit{StyleLipSync}, a style-based personalized lip-sync video generative model that can generate identity-agnostic lip-synchronizing video from arbitrary audio. To generate a video of arbitrary identities, we leverage expressive lip prior from the semantically rich latent space of a pre-trained StyleGAN, where we can also design a video consistency with a linear transformation. In contrast to the previous lip-sync methods, we introduce pose-aware masking that dynamically locates the mask to improve the naturalness over frames by utilizing a 3D parametric mesh predictor frame by frame. Moreover, we propose a few-shot lip-sync adaptation method for an arbitrary person by introducing a sync regularizer that preserves lip-sync generalization while enhancing the person-specific visual information. Extensive experiments demonstrate that our model can generate accurate lip-sync videos even with the zero-shot setting and enhance characteristics of an unseen face using a few seconds of target video through the proposed adaptation method. 

\end{abstract}

\section{Introduction}
\label{introduction}

In the past few years, advances in deep learning have altered the dynamics of video creation. Now, users can easily make and edit videos with the help of deep learning. In particular, the task of generating a talking head video has received great interest due to its various practical uses. It can be applied in many applications such as film dubbing into a different language, face-to-face live chats, and virtual avatars in games and videos. Thus, a lot of prior works \cite{lipgan, NeuralVP, wav2lip, makeittalk, pcavs, synctalkface} have been studied to generate a talking head video that has accurate lip shapes according to arbitrary audio inputs.

Most of the prior works mainly focus on enhancing synchronization between lip shapes and audio input. Some of the previous methods \cite{makeittalk, Das2020, NeuralVP} use intermediate structural representations such as landmarks and 3D models. They predicted the representations from the audio input and synthesized a talking head video of a target person. However, they suffered from inaccurate lip-sync results since such representations are too sparse to produce fine-grained details in lip-syncing. Recently, another line of methods \cite{wav2lip, synctalkface} mapped input audio to latent space and leveraged it to construct the mouth region of the target identity. While it achieves satisfactory results in lip-syncing, it generated blurry lower faces which are visually implausible. Furthermore, most methods only consider synthesizing frame-by-frame, lacking temporal consistency at the video level.

In this paper, we propose StyleLipSync, a style-based lip-sync video generative model that can generate identity-agnostic lip-synchronizing video from the arbitrary audio input. Our model consists of the following components. First, different from a previous masking method \cite{lipgan, wav2lip, synctalkface, videoretalking} which masks the entire lower half face, we propose Pose-aware Masking. We analyze that the previous masking method cause unpleasant artifacts and unnatural jaw moving in the generated videos. To circumvent this, we utilize a 3D face mesh predictor \cite{bfm, mediapipe} and generate lip masks with consideration of pose information and facial semantics such as jaw shape. Second, our image decoder is based on a style-based generator, namely StyleGAN \cite{stylegan1, stylegan2, stylegan3}. StyleGANs have demonstrated their effectiveness in various facial generative tasks, including face editing\cite{im2st, im2st++}, face enhancement \cite{gpen}, and video generation \cite{mocoganhd, styletalker}. As a pre-trained StyleGAN already contains expressive and diverse face priors in style latent space \cite{im2st}, we leverage it to synthesize the high-fidelity lip region of the target person. Furthermore, thanks to the continuous and linear nature of the latent space \cite{stylegan1, ganspace, sefa}, we linearly manipulate the style codes using the audio input to generate lip-synced video frames. Additionally, we propose Style-aware Masked Fusion to effectively adopt a skip-connection to our decoder, which helps to preserve the 2D structure of the image and improves lip fidelity. Finally, we propose a Moving-average based Latent Smoothing module that makes the latent trajectory smoother for enhancing the temporal consistency in the synthesized talking head video.

While our model can synthesize a talking head video of the target person, there is a slight identity gap between a generated video and the target person. The gap can be noticeable, for example, in racial faces, which are relatively scarce in the training data. One approach to addressing this issue is to fine-tune the generator on a few seconds of video of the target person to create a personalized model. Several fine-tuning methods \cite{pti, mystyle, stit} have already been demonstrated and widely adopted by the industry to achieve product-level quality. However, we analyze that simply fine-tuning the generator loses its ability to generalize to arbitrary audio inputs, which is critical for generating talking head videos. Therefore, to minimize the side effect, we propose a sync regularizer enforcing the audio generalization performance. The key idea is to leverage the audio from the training data, not from the target video. Specifically, we not only optimize the generator to reconstruct the target video but also synthesize the video corresponding to the randomly sampled audio from the training data and maintain a sync correlation between the synthesized video and the audio. As a result, we obtain a personalized lip-sync generative model that can synthesize a video of the target person for arbitrary audio.
Our contributions are summarized as follows:
\begin{itemize}
    \vspace*{-1.5mm}
    \item We present StyleLipSync, a lip-sync video generative model which generates lip-synchronizing video in-the-wild of $256\times256$ resolution with accurate and natural lip movement from a given masked video frames, audio segment, and single reference image.
    \vspace*{-2mm}
    \item We additionally propose a few-shot adaptation method for entirely unseen faces, which uses only a few seconds of video by introducing a sync regularizer to maintain audio generalization.
    \vspace*{-2mm}
    \item Experimental results demonstrate that StyleLipSync achieves state-of-the-art performance in terms of lip-sync and visual quality, even with the zero-shot setting.
\end{itemize}
\section{Related Works}
\label{sec:related}

\subsection{Lip-sync Video Generation}
\label{sec:lipsync}
Lip-sync video generation aims to generate a talking face video with lip motion synchronized with the given input audio. Early works \cite{lipgan, wav2lip} generate lip from the lower half masked face image corresponding to the input audio. Specifically, Wav2Lip \cite{wav2lip} uses a pre-trained SyncNet \cite{syncnet} as a lip-sync expert which maximizes the correlation between the generated lip and the input audio. Similarly, we use lip-sync expert for audio-visual alignment, which is trained with a contrastive manner proposed in \cite{styletalker}. VideoRetalking \cite{videoretalking} improves Wav2Lip \cite{wav2lip} a two-stage manner, where it first generates low-resolution ($96\times96$) video and then increases the resolution by using a single identity specific super-resolution network. In contrast, we propose a zero-shot model that directly generates lip-sync video of $256 \times 256$ resolution and also propose a unseen face adaptation method then enhances the personal characteristics from a few-shot target video. SyncTalkFace \cite{synctalkface} introduces a lip memory network, which encodes lip motion features into a discrete space in the training phase and retrieves a lip feature from query audio at the inference phase. In contrast to \cite{synctalkface}, we utilize continuous and linear latent space from a pre-trained style-based generator \cite{stylegan2} to generate lip images with high fidelity and take video consistency into account in the latent space.

\subsection{GAN Prior}
\label{sec:prior}
Style-based generators \cite{stylegan1, stylegan2, stylegan3} demonstrate the power of their semantic latent spaces, namely $\w$, in image generation, image editing \cite{ganspace, sefa}, and video generation \cite{mocoganhd}. GAN-inversion \cite{gan, im2st, im2st++, psp, e4e} utilizes the pre-trained GANs to invert an image into corresponding latent code so one can manipulate the attributes of the image only within the latent space. Extended $\wplus$ have been shown its much expressive power. For instance, pSp \cite{psp} adopts the feature pyramid networks (FPNs) \cite{fpn} to use $\wplus$, which follow the nature of the progressive generation \cite{progan} of StyleGAN \cite{stylegan1} and achieves state-of-the-art performance in image-to-image translation (e.g., face in-painting). Similarly, we utilize $\wplus$ for diverse and strong lip prior since we aim to build a lip-sync video generative model of arbitrary identity. 

Recent works \cite{gpen, gfpgan, styleswap} not only adopt pre-trained GAN prior as their decoder but also introduce a skip-connection that concatenates the encoded and generated features, which helps the model preserve 2D spatial information. Specifically, GPEN \cite{gpen} uses skip-connection that concatenates the encoded and generated features and achieves state-of-the-art performance in blind face restoration. StyleSwap \cite{styleswap} adopts it to face swap and introduces the ToMask branch that predicts the target facial attribute regions for swapping in a supervised manner. In contrast to those methods, we use an additive skip connection more efficient than the concatenation, along with the unsupervised predicted masked sum, which helps the decoder distinguish the target lip region from the whole face and therefore increases the lip image fidelity.

\subsection{Personalization}
\label{sec:personalization}
\begin{figure*}[t]
\begin{center}
\includegraphics[width=0.95\linewidth]{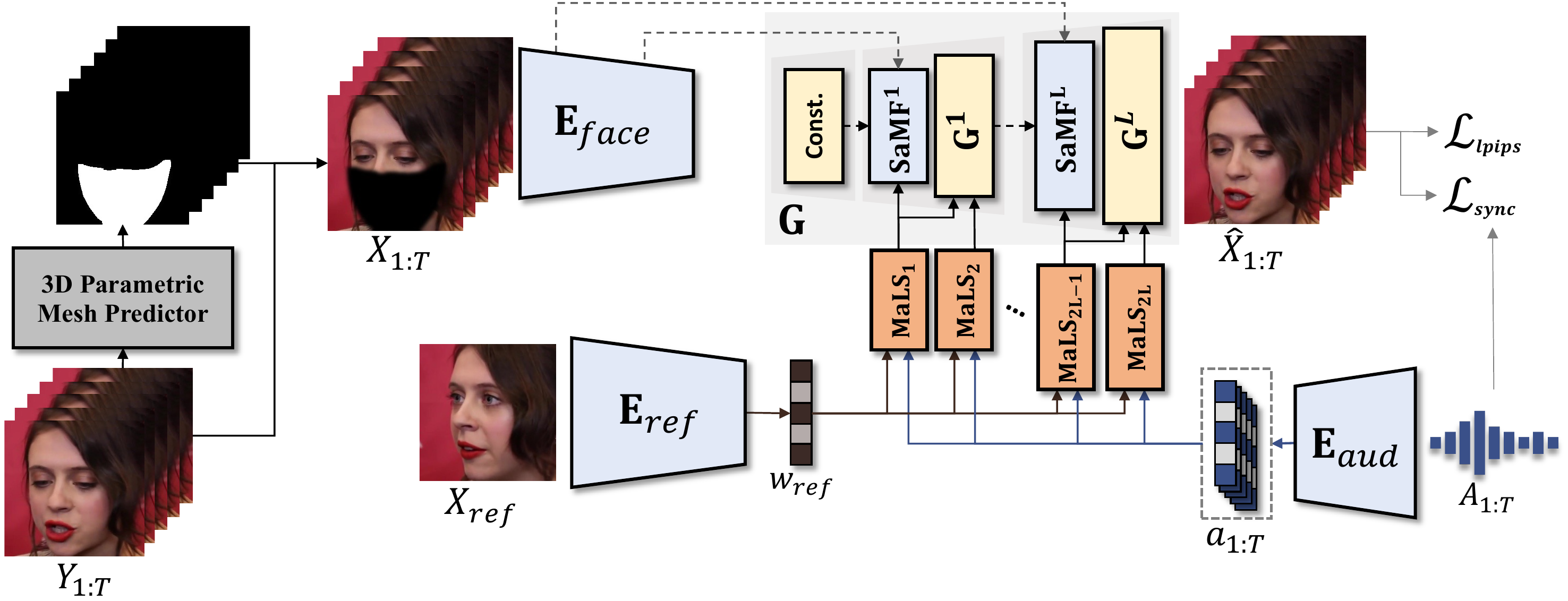}
\end{center}
\vspace*{-4mm}
   \caption{A framework of StyleLipSync. We leverage a 3D parametric mesh predictor \cite{bfm, mediapipe} to obtain pose-aware masked frames $X_{1:T}$, which inherits the facial pose of input frames. Face encoder $\faceencoder$ maps $X_{1:T}$ into 2D spatial features and then fed into the decoder $\decoder$ through \textit{style-aware masked fusion} ($\samf$). Single reference image $X_{ref}$ and audio segments $A_{1:T}$ are mapped into latent space, followed by \textit{Moving-average based Latent Smoothing} ($\mals$). This module outputs smooth video latent codes $\wsmooth_{1:T} \subseteq \wplus$ that represent temporally consistent lip movement. With the guidance of $\samf$s and the smooth video latent codes $\wsmooth_{1:T}$, StyleLipSync can generate temporally consistent lip-synced videos.}
\label{fig:overview}
\end{figure*}
Although GAN prior has been successful in various tasks \cite{psp, e4e}, it is still challenging to faithfully recover person-specific information that lies in the out-of-distribution \cite{im2st, im2st++, psp, e4e, gpen}. Recently, a few-shot personalization has become an alternative to solving the problem \cite{mystyle}. Pivotal-tuning-inversion (PTI) \cite{pti} fine-tunes the image generator while freezing a single latent code, namely \textit{pivot}, to compensate the person-specific information in the generative process, not in the encoding process. MyStyle \cite{mystyle} adopts PTI \cite{pti} to image inpainting and semantic editing, by restricting the latent space to a subspace spanned by the multiple pivots from the few photos (roughly 100) of the target person. Stitch-it-in-time \cite{stit} adopts the pivotal-tuning to the video editing in a multi-stage manner, which leverages an off-the-shelf latent manipulation \cite{interfacegan} to manipulate video in the latent space and then stitch it to the source video. Inspired by them, we propose a few-shot unseen face adaptation method that slightly fine-tunes the image decoder for a given latent code trajectory of a target identity and maintains the audio generalization by introducing a sync-regularizer.

\section{Method}
\label{sec:method}

Given lip masked video frames $X_{1:T} = (X_t)_{t=1}^{T}$ and audio segments $A_{1:T} = (A_t)_{t=1}^{T}$, a lip-sync method generates video frames 
\begin{equation} 
    \hat{X}_{1:T} = (\hat{X}_t)_{t=1}^{T},
    \label{eq:x_hat}
\end{equation}
where $\hat{X}_{1:T}$ has a lip movement synchronized with the audio segments $A_{1:T}$. In contrast to the previous lip-sync methods \cite{lipgan, wav2lip, synctalkface, videoretalking}, we leverage 3D parametric facial mesh predictor \cite{bfm, mediapipe} to compute lip mask so that the generator can be aware of semantically meaningful facial pose information (Section \ref{sec:poseaware}). We utilize a pre-trained StyleGAN \cite{stylegan2} as our decoder $\decoder$. When audio encoder $\audencoder$ and reference encoder $\refencoder$ map their inputs into the latent space $\wplus$ (Section \ref{sec:encoders}), the decoder $\decoder$ generates lip-synced video frames $\hat{X}_{1:T}$ from these latent codes (Section \ref{sec:decoder}) guided by the proposed \textit{style-aware masked fusion}. For enhancing the temporal consistency, we propose a \textit{Moving-average based Latent Smoothing} module,  which learns local motion between the latent codes, and makes video latent trajectory smoother. Finally, sync loss \cite{syncnet, styletalker} is used for the audio-lip synchronization. The overall framework of our model is described in Figure \ref{fig:overview}.

\subsection{Pose-aware Masking}
\label{sec:poseaware}
\begin{figure}[t]
\begin{center}
    \includegraphics[width=\linewidth]{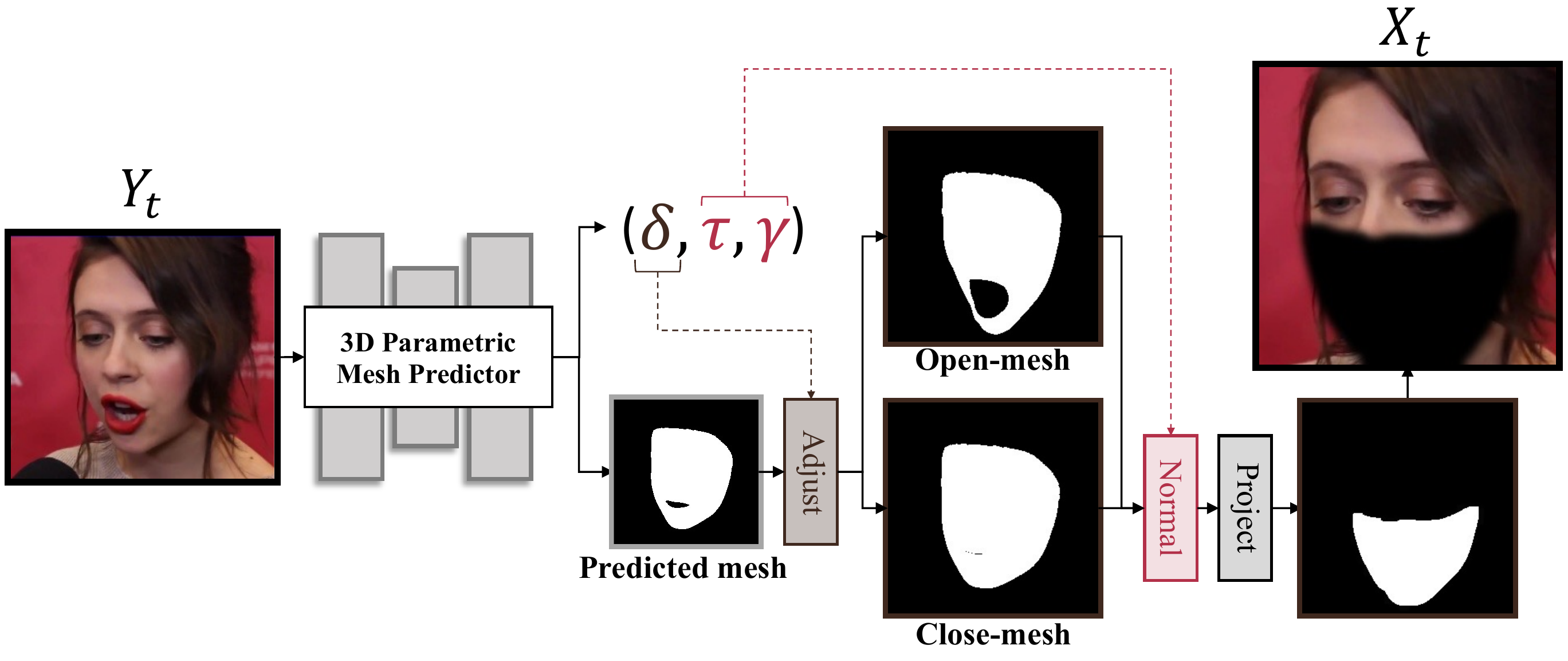}
\end{center}
\vspace*{-4.6mm}
\caption{Illustration of pose-aware masking. The expression parameter $\delta \in \real^{64}$ and the pose parameter $\tau \in \real^{3}, \gamma \in \text{SO}(3)$ are used to compute the natural mask.}
\label{fig:masking}
\vspace*{-4.6mm}
\end{figure}

Dynamic head motion is an important factor in the natural talking style. However, existing methods \cite{lipgan, wav2lip, synctalkface, videoretalking} employ the rectangular lower-half mouth masking method without consideration of the pose information. It often fails to detect appropriate masking regions when the head moves dynamically, which leads to unpleasant artifacts and unnatural jaw movement in the generated videos (see the first row in Figure \ref{fig:reconstruction} for examples).
To address this limitation, we use the face meshes by leveraging a 3D face mesh predictor \cite{bfm}, which captures 3D parameters and predicts dense face geometry. We predict the 3D parameters \cite{3dmm} and the mesh from given video frames. Then, the predicted expression parameter $\delta \in \real^{64}$ is used to adjust the mesh to obtain naturally opened and closed mouth meshes. We normalize the mesh vertices using the predicted pose parameters $\tau \in \real^{3}$ (translation), $\gamma \in \text{SO}(3)$ (rotation) and leave only the lower frontal vertices. These meshes are combined and projected onto the original 2D plane to finally get our pose-aware lip masks. Figure \ref{fig:masking} illustrates the framework of the pose-aware masking. This masking not only captures the pose information but also inherits facial semantics such as jaw shape. Ablation studies in Section \ref{sec:ablation} show that the pose-aware masking helps the model to increase visual quality along with dynamic pose.

\subsection{Decoder}
\label{sec:decoder}
\noindent \textbf{Lip Prior from Style-based Generator.} \indent Generating the lip-sync videos from scratch for an arbitrary person is very hard since the mapping from audio to lip is basically one-to-many. In this paper, we leverage a style-based generator as our image decoder $\decoder$ for the following two reasons. First, a pre-trained StyleGAN already contains expressive and diverse face priors \cite{stylegan1, stylegan2, stylegan3} in the form of latent code, namely \textit{style code} in the latent space  $\wplus$ \cite{im2st++}. Those latent spaces enable us to better synthesize the lip region of the target person with the diverse lip prior. Second, the style codes form the continuous and linear \cite{stylegan1, ganspace, sefa} latent spaces, which enables us to design a high-level visual transformation, such as natural motion, only with a linear transformation of latent code \cite{lia}. Hence, we can generate a talking head video with smooth lip motion by simply manipulating the style codes using audio, which the previous lip-sync methods never take into account.

\noindent \textbf{Style-aware Masked Fusion ($\samf$).} \indent Recently, it has been explored that adopting skip-connections based on the concatenation to GAN-inversion helps to preserve the 2D spatial information of the input \cite{styleswap, gcfsr, gpen}. Similarly, we adopt an additive skip-connection to our model for effectively preserving the non-masked region of $X_{1:T}$ while faithfully utilizing the latent space.

\begin{figure}[t]
\begin{center}
    \includegraphics[width=\linewidth]{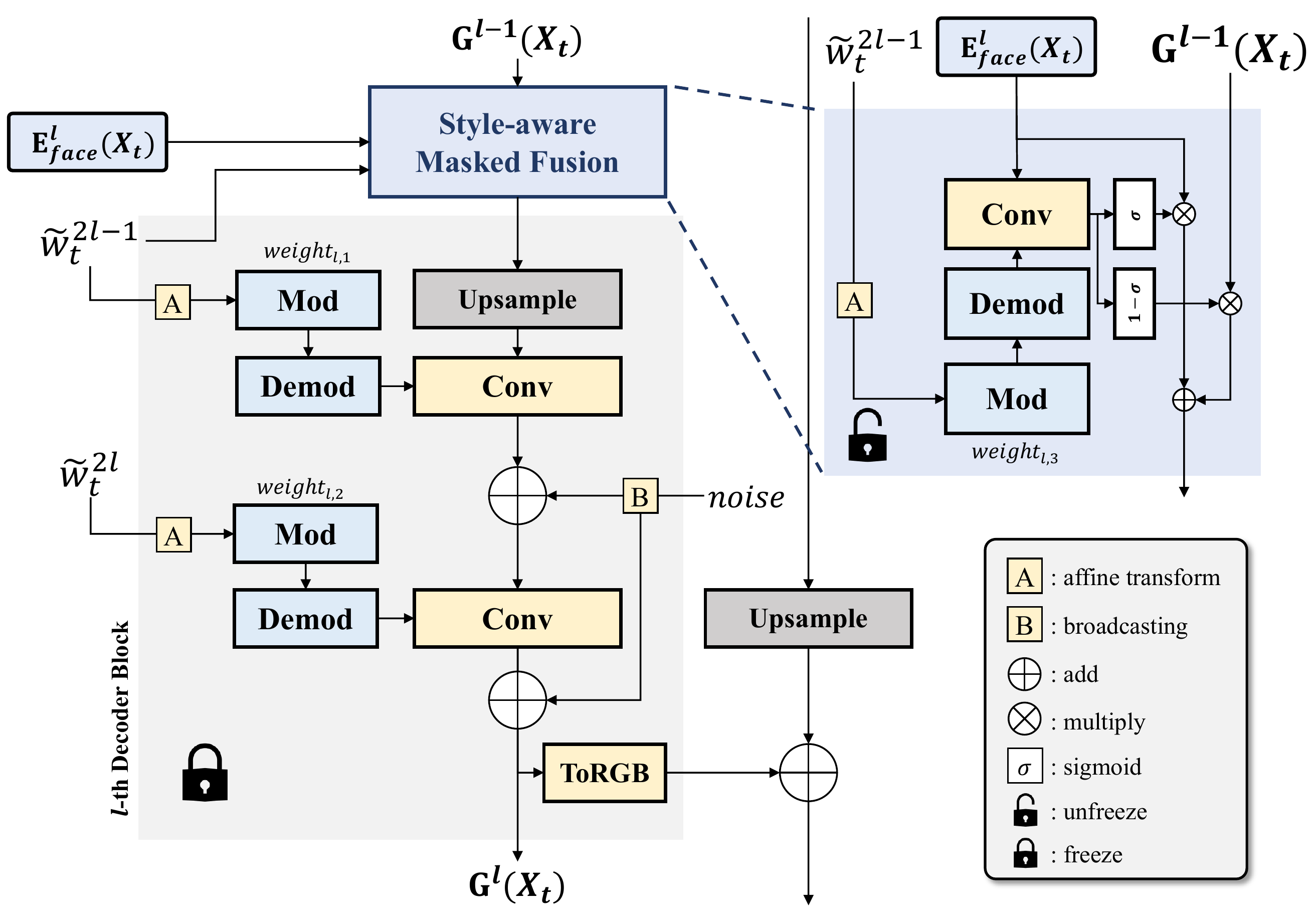}
\end{center}
\vspace*{-4mm}
\caption{Illustration of the decoder block. The encoded feature $\faceencoder^l(X_t)$ is injected into $l$-th decoder block through Style-aware Masked Fusion ($\samf$). Note that only the convolutions in $\samf$ are trainable, while the others are frozen during the training phase.}
\label{fig:decoder}
\vspace*{-4mm}
\end{figure}

Specifically, we propose \textit{style-aware masked fusion} ($\samf$) for efficiently preserving the 2D spatial feature and relieving the information gap between masked and non-masked regions. $\samf$s are introduced at the beginning of the decoder blocks. The decoder $\decoder$ consists of $L$ decoder blocks, each of which takes 2 style codes to modulate 3 convolution weights, as illustrated in Figure \ref{fig:decoder}. The first style code in each decoder block modulates 2 different convolution weights, one for the convolution in the original block and the other for the $\samf$. $\samf$ learns to predict a 1-channel mask $S^{l}_t$ of the current resolution from the encoded feature through the newly modulated convolution followed by the sigmoid, which is used for spatial weighted fusion of the encoded feature and the generated feature.

Formally, given encoded feature $\faceencoder^{l}(X_t)$ and generated feature $\decoder^{l-1}(X_t)$ of same dimension $\real^{h \times w \times c}$, $\samf$ firstly predicts a spatial mask $S^{l}_{t} \in \real^{h \times w \times 1}$ from $\faceencoder^{l}(X_t)$ and then output fused feature as follows:
\begin{equation}
    S^{l}_{t} \otimes \faceencoder^{l}(X_t) + (1-S^{l}_{t}) \otimes \decoder^{l-1}(X_t).
    \label{eq:fusion}
\end{equation}
Ablation studies (Section \ref{sec:ablation}) show that $\samf$s improve the fidelity of the mouth since they separate masked and non-masked regions.

\subsection{Encoders}
\label{sec:encoders}
Our model has three different encoders: face encoder $\faceencoder$, reference encoder $\refencoder$, and audio encoder $\audencoder$.

Face encoder $\faceencoder$ takes masked video frames $X_{1:T}$ as input and then outputs $l$ 2D spatial features $ \faceencoder^{l}(X_t) = \{\faceencoder^l(X_t) ~|~ l \in [1, 2, \cdots, L]\}$ for each $t$. These features are injected into the decoder $\decoder$ through the style-aware masked fusion to efficiently preserve 2D spatial structure, as described in Section \ref{sec:decoder}.

Reference Encoder $\refencoder$ maps a reference $X_{ref}$ into $2L$ reference style codes, each of which has 512 dimensions. We simply denote the reference style code as $w_{ref} = [w^1_{ref} | w^2_{ref} | \cdots | w^{2L}_{ref}] \in \real^{512 \times 2L}$.

Similar to $\refencoder$, audio encoder $\audencoder$ maps a single audio segment into $2L$ audio style codes, each of which has 512 dimensions. As we use $T$ consecutive audio segments $A_{1:T}$, $\audencoder$ independently maps $A_{t}$ into $a_{t} = [a^{1}_{t} | a^{2}_{t} | \cdots | a^{2L}_{t}] \in \real^{512 \times 2L}$. We simply denote $a_{1:T} = (a_t)_{t=1}^{T}$ for $T$ audio style codes. Please refer to our supplementary materials for the detailed encoder architectures.

From these style codes $w_{ref}$ and $a_{1:T}$, we compute target video's style codes over frames $w_{1:T}$ by
\begin{equation}
    w_{1:T} = (w_1, w_2, \cdots, w_T), 
    \label{eq:wt}
\end{equation}
where $w_{t} = w_{ref} + a_{t} \in \wplus$. We compute the style codes by simply adding these two different codes based on the linearity \cite{ganspace, sefa} of the latent space $\wplus$. The style codes $w_{1:T}$ are then fed into the decoder $\decoder$ through the learned affine transformations to generate synced lip motion.

\noindent \textbf{Temporal Consistency.} \indent Thanks to the semantically rich latent space $\wplus$, our model can generate an accurate lip-sync video frame by frame. However, this frame-wisely independent encoding of style codes turns out to lead inconsistent mouth movements in the final results. To remedy this, we assume that the generated style codes $w_{1:T}$ form a trajectory of the target video \cite{dynode} in $\wplus$ and enforce the smooth local transition of the motion \cite{styletalker, fomm, lia} into the trajectory.
Toward this, we introduce \textit{Moving-average based Latent Smoothing} ($\mals$), each of which consists of a stack of the weighted moving-average \cite{thirdtime} and 1D convolutions operating on the style codes along the time-axis. More precisely, we employ $2L$ $\mals$ for $L$ resolutions, each of which takes the $l$-th component of $w_{t-1}, w_t$, and $w_{t+1}$ as its inputs to learn the local difference between them, and then we inject the local motions into $w_{ref}$ to compute the smooth style code $\wsmooth_t$:
\begin{eqnarray}
    \wsmooth^{l}_{t} & = & w^{l}_{{ref}} + \mals^{l}(w^l_{t-1:t+1}), \nonumber \\
                     & = & w^{l}_{{ref}} + \text{Conv1Ds}\left(\sum_{\tau=t-1}^{t+1} \alpha_{\tau}\cdot w^l_{\tau}\right),\label{eq:mals}
\end{eqnarray}
where $\mals^{l}$ denotes the $l$-th $\mals$, and $\alpha_{\tau}$ is the weight of the moving-average.

With this smooth latent codes $\wsmooth_{1:T}$, we compute the video frames:
\begin{equation}
    \hat{X}_{1:T} =  \left( \decoder(\wsmooth_t, \faceencoder(X_t)) \right)_{t=1}^{T}.
\label{eq:video}
\end{equation}
For better initialization \cite{psp}, we add the average code $w_{avg}$ of the pre-trained generator to each $\wsmooth_{t}$ in \eqref{eq:mals}. 

\subsection{Training Objective}
\label{sec:train}
We train StyleLipSync to reconstruct target video frames from corresponding audio. We randomly choose $T=5$ consequent frames with corresponding audio segments and 1 single reference frame. Image perceptual loss \cite{lpips} is used to minimize perceptual image distance between generated frames $\hat{X}$ and ground-truth frames $Y$:
    \begin{equation}
        \loss_{lpips} = \sum_{i=1}^{N} \left\| \phi^{i}(\hat{X}) - \phi^{i}(Y) \right\|_2,
        \label{eq:lpips}
    \end{equation}
where $N$ is the number of feature extractor, $\phi^{i}(\cdot)$ is the $i$ th feature extractor, and $\| \cdot \|_{2}$ is the $\ell^{2}$ loss. Similar to \cite{fomm}, we use the multi-scale perceptual loss with 3 levels.

For audio-visual alignment, we utilize SyncNet trained in a contrastive manner \cite{styletalker} that minimizes the cosine distance between generated frames $\hat{X}$ and corresponding audio segment $A$:
    \begin{equation}
        \loss_{sync} = 1 - \cos(f_v (\hat{X}), f_a(A)), 
        \label{eq:sync}
    \end{equation}
where $\cos(\cdot, \cdot)$ denotes the cosine similarity. $f_{v}(\cdot)$ and $f_{a}(\cdot)$ are the frame and audio feature extractor, respectively. The final objective $\loss_{train}$ is computed as:
\begin{equation}
    \loss_{train} = \lambda_{1} \loss_{lpips} + \lambda_{2} \loss_{sync},
    \label{eq:train}
\end{equation}
where $\lambda_1$ and $\lambda_2$ are the balancing coefficients.
\section{Unseen Face Adaptation}
\label{sec:adaptation}
\begin{figure}
\begin{center}
    \includegraphics[width=\linewidth]{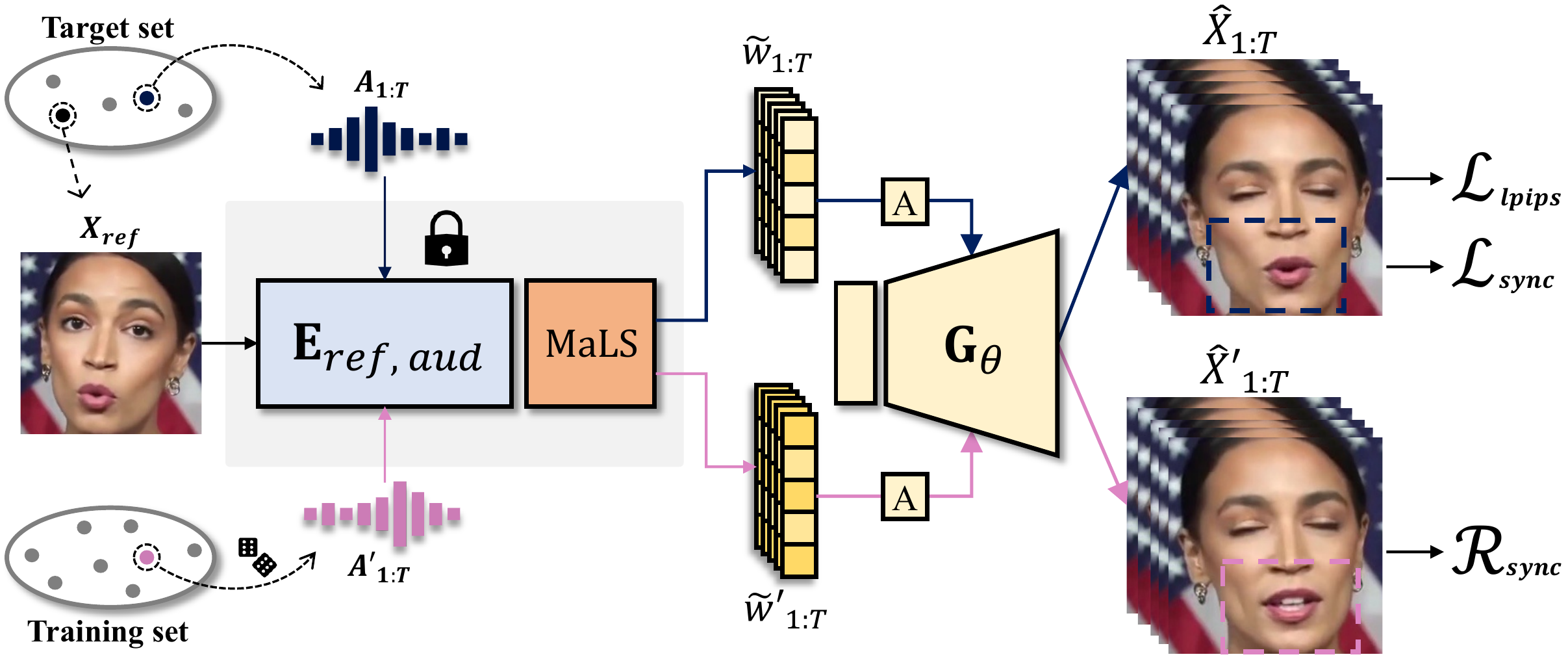}
\end{center}
    \vspace*{-4mm}
   \caption{Adaptation for Unseen Face. We slightly tune the decoder $\decoder_{\theta}$ with the proposed sync regularizer $\reg_{sync}$, while freezing all encoders' weight. Face encoder $\faceencoder$ and $\samf$s are omitted here for simplicity.}
\label{fig:adaptation}
\vspace*{-4mm}
\end{figure}
Although StyleLipSync successfully generates accurately lip-synced videos with high fidelity, the model would fail to exactly synthesize unseen faces lying in the out-of-distribution. This problem refers to the \textit{failure of id preservation}. Therefore, to handle this, we fine-tune our decoder $\decoder$ on a target person video to make a personal model that is better able to synthesize toward the target person.


Let $X_{1:T}$ be masked video frames of \textit{unseen face}, which corresponds to the audio segments $A_{1:T}$, and $X_{ref}$ be a reference frame. The frozen encoders convert each input into the intermediate representations, $\wsmooth_{1:T}$ and $\faceencoder(X_{t})))_{t=1}^{T}$. Then we fine-tune the decoder $\decoder_{\theta}$, now parameterized by $\theta$, by minimizing the distance between $(\decoder_{\theta}(\wsmooth_{t}, \faceencoder(X_{t})))_{t=1}^{T}$ and target frames, as same as \eqref{eq:train}.
\begin{figure*}[!tb]
\begin{center}
    \subfigure[Reconstruction results on Voxceleb2 \cite{vox2} test data.]{\includegraphics[width=0.495\textwidth]{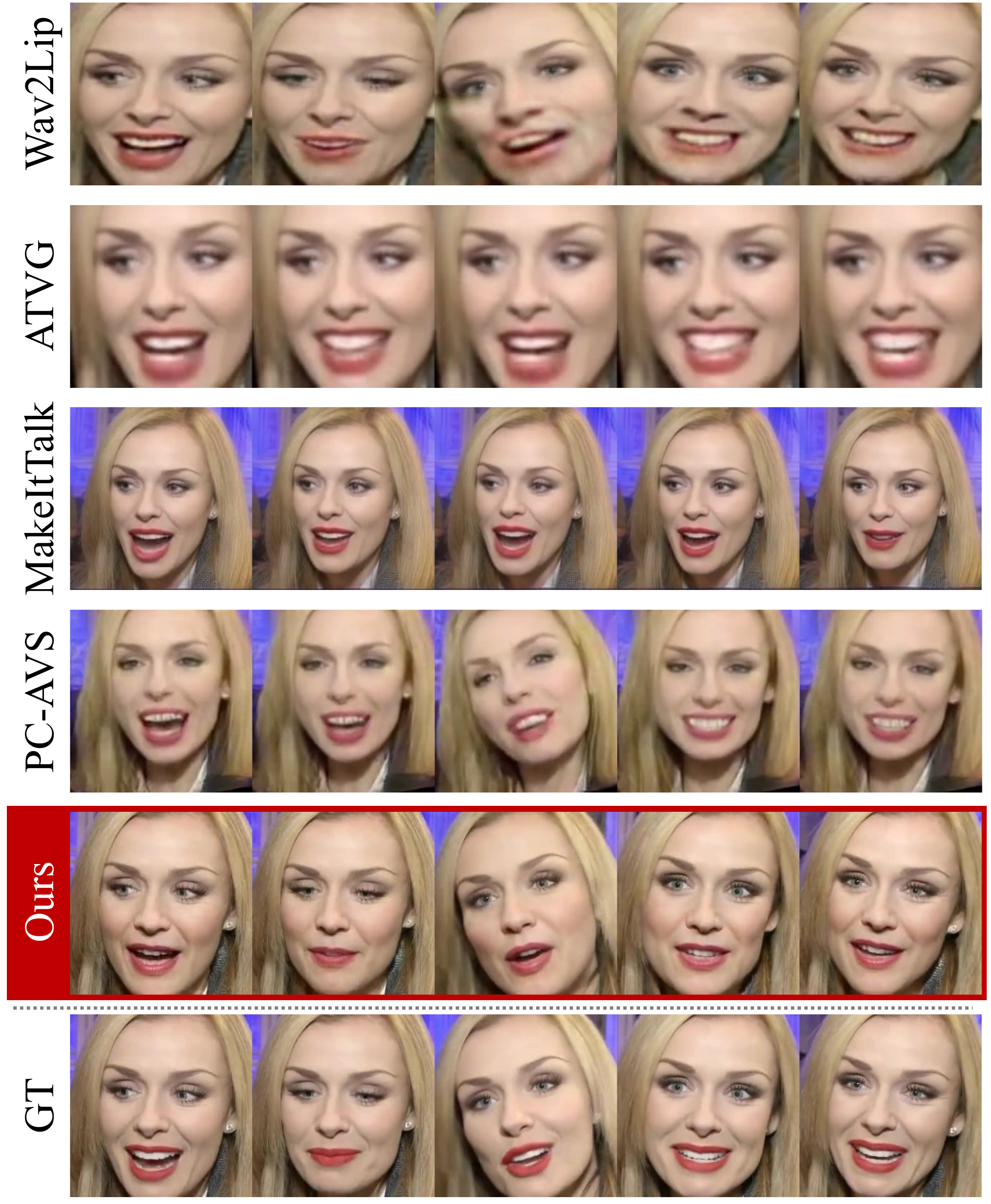}}\label{fig:compare_reconstruction}
    \subfigure[Cross-id results on HDTF \cite{hdtf}.]{\includegraphics[width=0.495\textwidth]{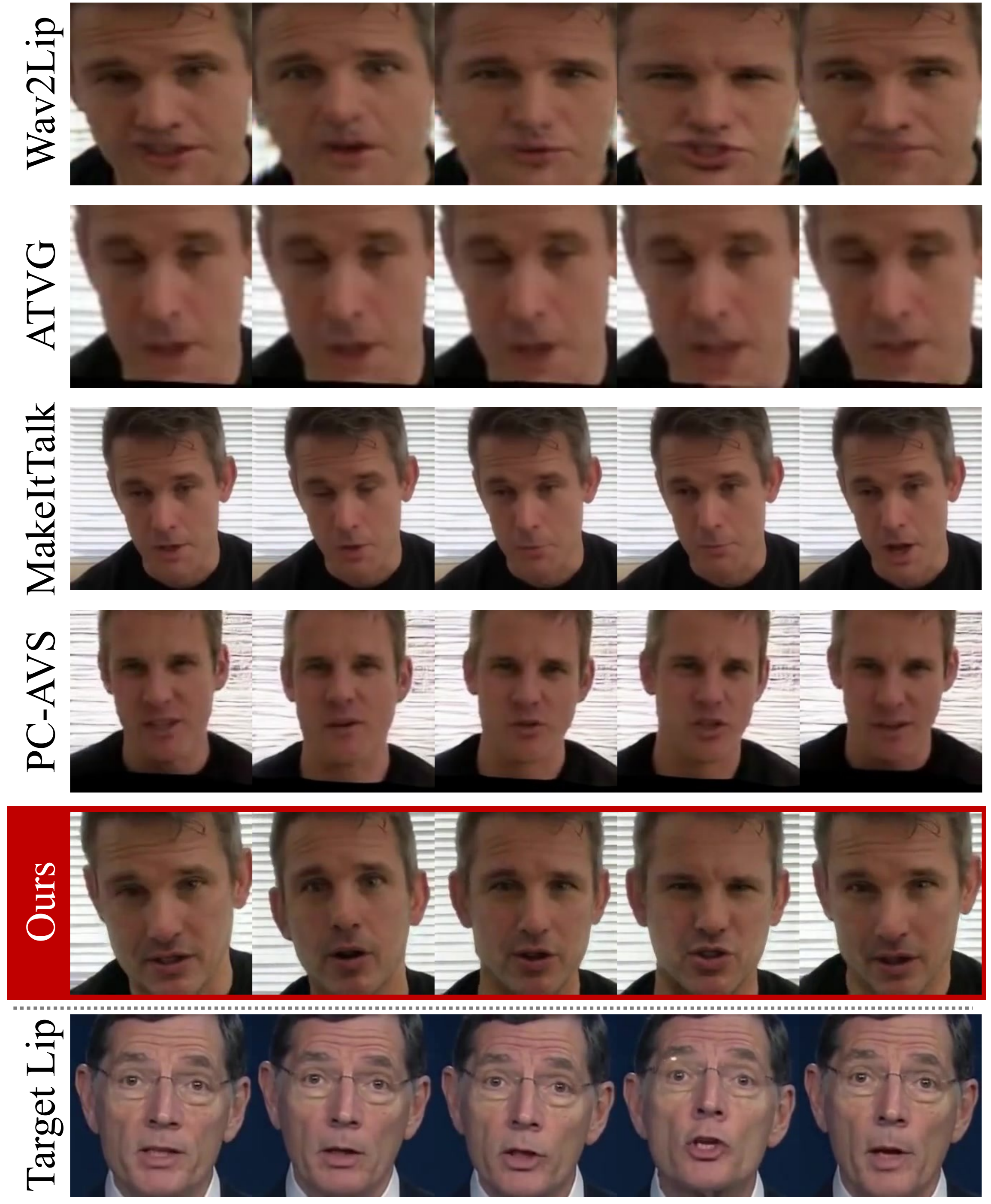}}\label{fig:compare_crossid}
\end{center}
\vspace*{-3mm}
   \caption{Comparison with state-of-the-art methods. The different field of view comes from the pre-processing strategy of each model.}\label{fig:reconstruction}
\vspace*{-3mm}
\end{figure*}
However, fine-tuning the decoder on a short video of a single identity leads to over-fitting and losing the lip-sync generality as the generator can memorize the target video \cite{fewshotcrossdomain}. To prevent the model from this scenario, we introduce a sync regularizer $\reg_{sync}$ to enforce audio generality to the decoder $\decoder_{\theta}$ by leveraging the audio from the training dataset, not from the target video. Formally, given audio segments $A'_{1:T}$ randomly chosen from the training data (Voxceleb2 \cite{vox2}) and $w_{ref}$, we compute smooth style codes $\wsmooth'_{1:T}$, and then decode them to a synced video $X'_{1:T}$. The sync regularizer $\reg_{sync}$ is defined as
\begin{equation}
    \reg_{sync} = 1 - \cos(f_v(\hat{X}'), f_a(A')),
    \label{eq:syncreg}
\end{equation}
which enforces $\decoder_{\theta}$ to generate $\hat{X}'_{1:T}$ aligning with $A'_{1:T}$. The final object for a single person adaptation is given as follows:
\begin{equation}
    \theta^{*} = \underset{\theta}{\mathrm{argmin}} ~ \loss_{train} + \lambda_R \reg_{sync},
    \label{eq:adap}
\end{equation}
where $\lambda_{R}$ is the regularizer coefficient. Ablation studies in Section \ref{sec:ablation} show that $\reg_{sync}$ regularizes the audio generality even if we use the audio from the training set.
\section{Experiments}
\label{sec:experiments}

\subsection{Dataset}
\label{sec:dataset}

We train our model on Voxceleb2 \cite{vox2}, which consists of in-the-wild talking face videos collected from YouTube. It contains more than 145,000 videos of about 6,100 identities in the train set and more than 4,900 videos of more than 110 identities in the test set. We convert all videos into frames with 25 fps and then crop and resize those frames into $256 \times 256$ resolution, following the method of \cite{fomm}. For more semantically rich face priors, we only use videos where the detected face bounding boxes' height (and width) is longer than 256. After pre-processing, the remaining 11051 videos and 340 videos are used as the training set and the test set, respectively. All audios are re-sampled to 16kHz and converted into a mel-spectrogram to be used as our audio representation similar to the method in \cite{pcavs, styletalker}. We also use HDTF \cite{hdtf} to further test our model for cross-id experiments. It is widely used to evaluate high-resolution talking face generative models, where the head pose dynamics are not as significant as the Voxceleb2.
\subsection{Implementation Details}
\label{sec:implementation}
We pre-train StyleGAN2 \cite{stylegan2} on the Voxceleb2 \cite{vox2} following the implementation of \cite{stylegan-ada}. For training StyleLipSync and SyncNet, we set the frame length $T=5$, and employ Adam \cite{adam} optimizer with learning rate $10^{-4}$ throughout the training phase in both cases. All experiments are performed on 2 TITAN RTX GPUs. We set $\lambda_1 = 10, \lambda_2 = 0.1$, $\lambda_{R} = 0.1$ for the objective \eqref{eq:adap}, and $\alpha_{t-1}=0.25$, $\alpha_{t}=0.5$, $\alpha_{t+1}=0.25$ for $\mals$ \eqref{eq:mals}. For all inference, we use the first frame as the reference image.

For image quality metric, we use PSNR, and SSIM. We also calculate CSIM \cite{arcface} in the cross-id experiment that measures the face similarity between the images in the pre-trained face embedding space. For lip-sync quality metric, we use LMD, LSE-D, and LSE-C. LMD \cite{atvg} is the absolute distance of facial landmarks between the target and generated frames. LSE-D and LSE-C are proposed in \cite{wav2lip}, where LSE-D measures the distance between the lip and audio representations, and LSE-C measures the lip-sync confidence, respectively.

\subsection{Evaluation}
\noindent \textbf{Reconstruction Results.} \indent We compare the other state-of-the-art methods in the lip-sync (Wav2Lip \cite{wav2lip}) and talking face generation (ATVG \cite{atvg}, MakeItTalk \cite{makeittalk}, and PC-AVS \cite{pcavs}) on reconstruction results of Voxceleb2 test dataset. Table \ref{tab:reconstruction} shows that StyleLipSync outperforms all metrics except the LSE-C score. Wav2Lip \cite{wav2lip} achieves the highest LSE-C score, however, it achieves low image quality metrics since it generates $96\times96$ low-resolution videos. PC-AVS \cite{pcavs} achieves comparable lip-sync scores with StyleLipSync, however, achieves low image quality metrics than our model since it highly relies on its specific pre-processing and fails to generate in many cases. We also illustrate the qualitative results in Figure \ref{fig:compare_reconstruction}. Wav2Lip \cite{wav2lip} produces a lip-synchronized video with considerable visual artifacts since it cannot adapt to the video of dynamic head pose. MakeItTalk \cite{makeittalk} generates low lip-sync video since it uses sparse facial landmarks. PC-AVS \cite{pcavs} generates an accurate lip-synchronized video following the input head pose. However, it struggles to preserve the unseen identity involving visual artifacts. StyleLipSync generates the natural lip motion with high fidelity and preserves the input identity, which is comparable to the ground truth video.

\noindent \textbf{Cross Identity Results.} \indent To evaluate lip-sync generalization, we conduct cross-id experiments settings using unseen videos and audio. We randomly sample 10 videos of different identities and 150 audio without duplication from HDTF \cite{hdtf}. We use the first 10 seconds of videos and audio. For each video, we generate 15 lip-synced videos from 15 different audios, where those face-audio pairs are not from the same source. In Table \ref{tab:cross}, we report the LSE-D and LSE-C for lip-sync quality and the CSIM for face similarity. Wav2Lip \cite{wav2lip} achieves a higher LSE-C score than StyleLipSync, however, it achieves a low CSIM score since it produces a low-resolution video with visual artifacts. MakeItTalk \cite{makeittalk} achieves the best CSIM score, while the lip-sync quality is the worst since it uses sparse facial landmarks. PC-AVS \cite{pcavs} outperforms LSE-C while achieving the lowest CSIM since it can't preserve an unseen face's identity. StyleLipSync achieves the best LSE-D score and comparable CSIM score to MakeItTalk \cite{makeittalk}. We show qualitative results of cross-id experiments with target lip references in Figure \ref{fig:compare_crossid}. PC-AVS \cite{pcavs} can generate accurate lip-synchronizing video compared to the target lip, while it fails to preserve facial details of the unseen face. MakeItTalk \cite{makeittalk} produces a high-resolution and identity-preserving video, however, it is out-of-sync compared to the target. StyleLipSync generates a high-resolution lip synchronizing video compared to the target lip, without any visual artifacts.
\begin{table}[tb]
    \caption{Quantitative comparison of reconstruction on Voxceleb2 test data. The best score for each metric is in \textbf{bold}.}
    \label{tab:reconstruction}
    \vspace*{-4.5mm}
    \begin{center}
    \scalebox{0.7}{
    \begin{tabular}{l || c | c | c | c | c}
    \toprule
        \multicolumn{1}{c||}{} & \multicolumn{5}{c}{Voxceleb2 (Reconstruction)}\\
        \multicolumn{1}{l||}{Method} & \multicolumn{2}{c}{Image} & \multicolumn{3}{c}{Lip-Sync} \\
        & SSIM $\uparrow$ & PSNR $\uparrow$ & LMD $\downarrow$ & LSE-D $\downarrow$ & LSE-C $\uparrow$\\
        \hline
        Wav2Lip$_{96\times96}$ \cite{wav2lip} & 0.448 & 13.534 & 6.422 & 6.999 & \textbf{8.329} \\
        ATVG$_{128\times128}$ \cite{atvg} & 0.461 & 13.349 & 7.165 & 8.821 & 5.421 \\
        MakeItTalk$_{256\times256}$ \cite{makeittalk} & 0.419 & 12.868 & 3.649 & 10.895 & 3.624 \\
        PC-AVS$_{224\times224}$ \cite{pcavs} & 0.369  & 13.210 & 2.812 & 7.278 & 7.699 \\ 
        \cdashline{1-6}
        \textbf{Ours}$_{256\times256}$ & \textbf{0.631} & \textbf{19.607} & \textbf{2.696} & \textbf{6.628} & 8.056\\
        \bottomrule
    \end{tabular}}
    \end{center}
    \vspace*{-4mm}
\end{table}
\begin{table}[tb]
    \caption{Quantitative comparison of cross-identity results for unseen face. We report CSIM \cite{arcface} as the image quality metric since there is no ground truth frames for the cross-id experiments. The best score for each metric is in \textbf{bold}.}\label{tab:cross}
    \begin{center}
        \scalebox{0.9}{
            \begin{tabular}{l||c|c|c}
                \toprule
                \multicolumn{1}{c||}{} & \multicolumn{3}{c}{HDTF (Cross-id)} \\
                \multicolumn{1}{l||}{Method} & \multicolumn{1}{c}{Image} & \multicolumn{2}{c}{Lip-Sync} \\
                & CSIM $\uparrow$ &LSE-D $\downarrow$ & LSE-C $\uparrow$ \\
                \hline
                Wav2Lip$_{96\times96}$ \cite{wav2lip} & 0.656 & 7.047 & 8.576  \\
                ATVG$_{128\times128}$ \cite{atvg} & 0.287 & 8.668 & 6.040 \\
                MakeItTalk$_{256\times256}$ \cite{makeittalk} & \textbf{0.770} & 10.641 & 4.725  \\
                PC-AVS$_{224\times224}$ \cite{pcavs} & 0.238 & 6.921 & \textbf{8.858}  \\
                \cdashline{1-4}
                \textbf{Ours}$_{256\times256}$ & 0.737 & \textbf{6.825} & 8.209 \\
                \bottomrule
            \end{tabular}
            }
    \end{center}
    \vspace*{-4mm}
\end{table}

\subsection{Ablation Studies}
\label{sec:ablation}
\begin{figure}[!tb]
\begin{center}
    \includegraphics[width=0.242\linewidth]{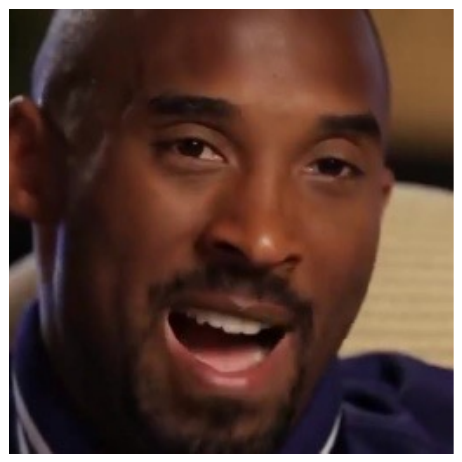}
    \includegraphics[width=0.242\linewidth]{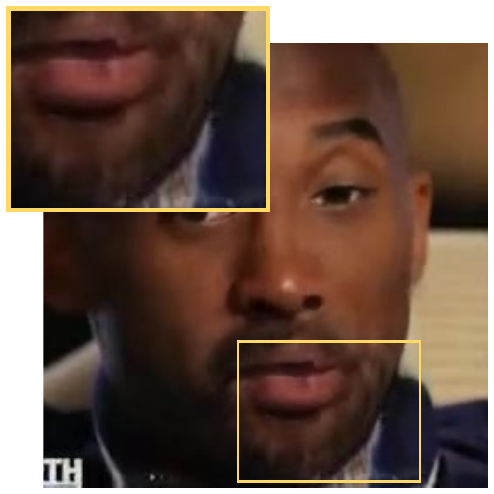}
    \includegraphics[width=0.242\linewidth]{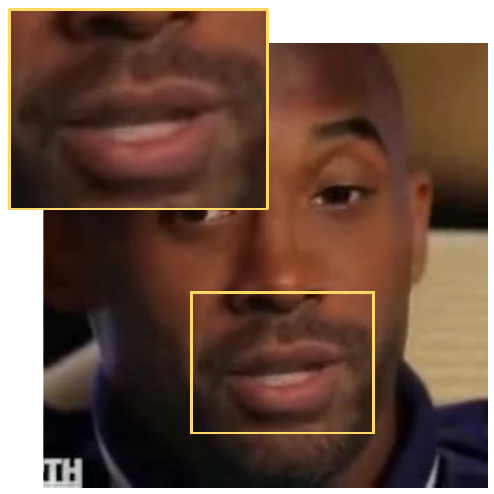}
    \includegraphics[width=0.242\linewidth]{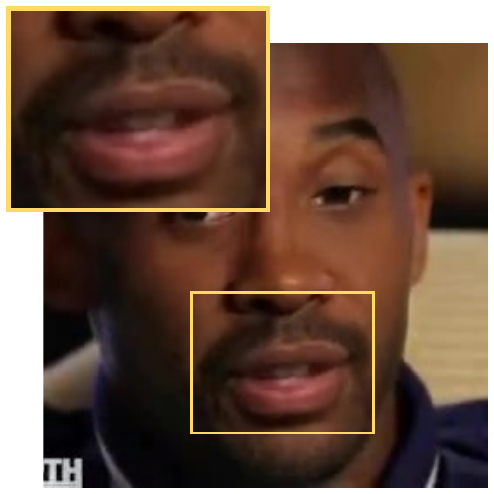}
\end{center}
\vspace*{-8.5mm}
\begin{center}
    \subfigure[\scriptsize{Reference}]{\includegraphics[width=0.242\linewidth]{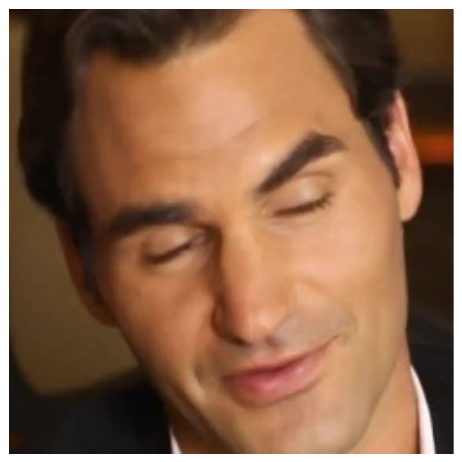}}
    \subfigure[\scriptsize{w/o Pose mask}]{\includegraphics[width=0.242\linewidth]{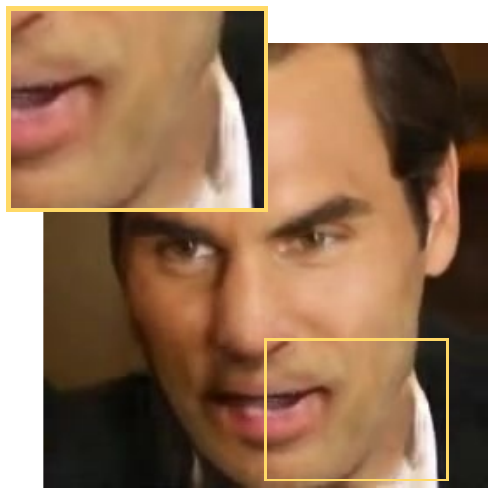}\label{fig:woPose}}
    \subfigure[\scriptsize{w/o $\samf$}]{\includegraphics[width=0.242\linewidth]{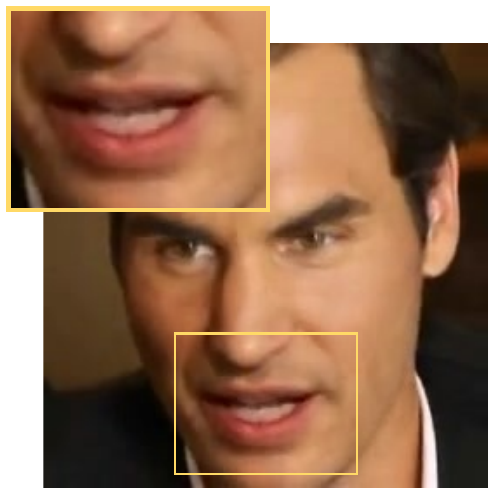}\label{fig:woSaMF}}
    \subfigure[\scriptsize{\textbf{Full}}]{\includegraphics[width=0.242\linewidth]{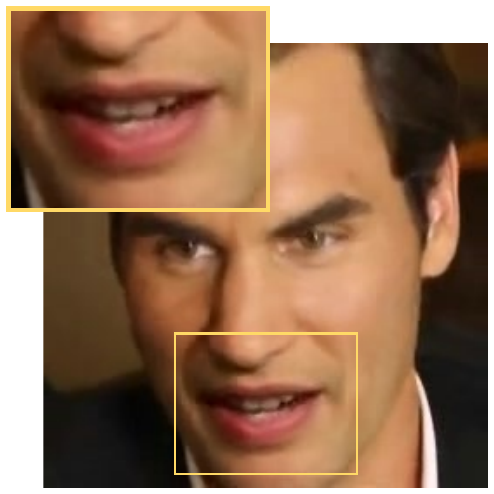}\label{fig:full}}
\end{center}
\vspace*{-4mm}
\caption{Qualitative comparison of zero-shot model.}
\label{fig:ablation_zeroshot}
\end{figure}

\noindent \textbf{Ablation Studies on Zero-shot Model.}  \indent
Figure \ref{fig:ablation_zeroshot} and Table \ref{tab:ablation_zeroshot} summarize the ablation studies on our zero-shot method of reconstruction on Voxceleb2 test data. If we replace the pose-aware masking with standard rectangular masking (w/o Pose Mask) in \cite{wav2lip, videoretalking, synctalkface}, considerable visual artifacts occur around the masked region since it is insufficient to capture the pose difference between the reference and the target. To validate $\samf$,  we replace the modulated convolutions in $\samf$s with the standard convolutions. Figure \ref{fig:woSaMF} shows that $\samf$s significantly improve lip region's fidelity since the modulated convolution helps to be aware of the lip style. As shown in Table \ref{tab:ablation_zeroshot}, $\mals$ significantly improves lip-sync quality, which cannot be reflected in a single image. Please refers to our project page for ablation studies on $\mals$.
\begin{table}[tb]
    \caption{Ablation study on zero-shot model. The best score for each metric is in \textbf{bold}.}\label{tab:ablation_zeroshot}
    \begin{center}
        \scalebox{0.78}{
            \begin{tabular}{l||c|c|c|c|c}
                \toprule
                \multicolumn{1}{c||}{} & \multicolumn{5}{c}{Voxceleb2 (Reconstruction)} \\
                \multicolumn{1}{l||}{Method (ours)} & \multicolumn{2}{c}{Image} & \multicolumn{3}{c}{Lip-Sync} \\
                & SSIM $\uparrow$ & PSNR $\uparrow$ & LMD $\downarrow$ & LSE-D $\downarrow$ & LSE-C \\
                \hline
                w/o Pose Mask & 0.602 & 18.867 & 3.057 & 6.771 & 7.748 \\
                w/o $\mals$  & 0.593 & 18.186 & 2.740 & 6.994 & 7.577  \\
                w/o $\samf$ & 0.591 & 18.181 & 2.764 & 6.838 & 7.780  \\
                \cdashline{1-6}
                \textbf{Full} & \textbf{0.631} & \textbf{19.607} & \textbf{2.696} & \textbf{6.628} & \textbf{8.056} \\
                \bottomrule
            \end{tabular}
            }
    \end{center}
    \vspace*{-4mm}
\end{table}

\begin{figure}[t]
\begin{center}
    \subfigure[Comparison of cross-id results of zero-shot and adaptation.]{\includegraphics[width=\linewidth]{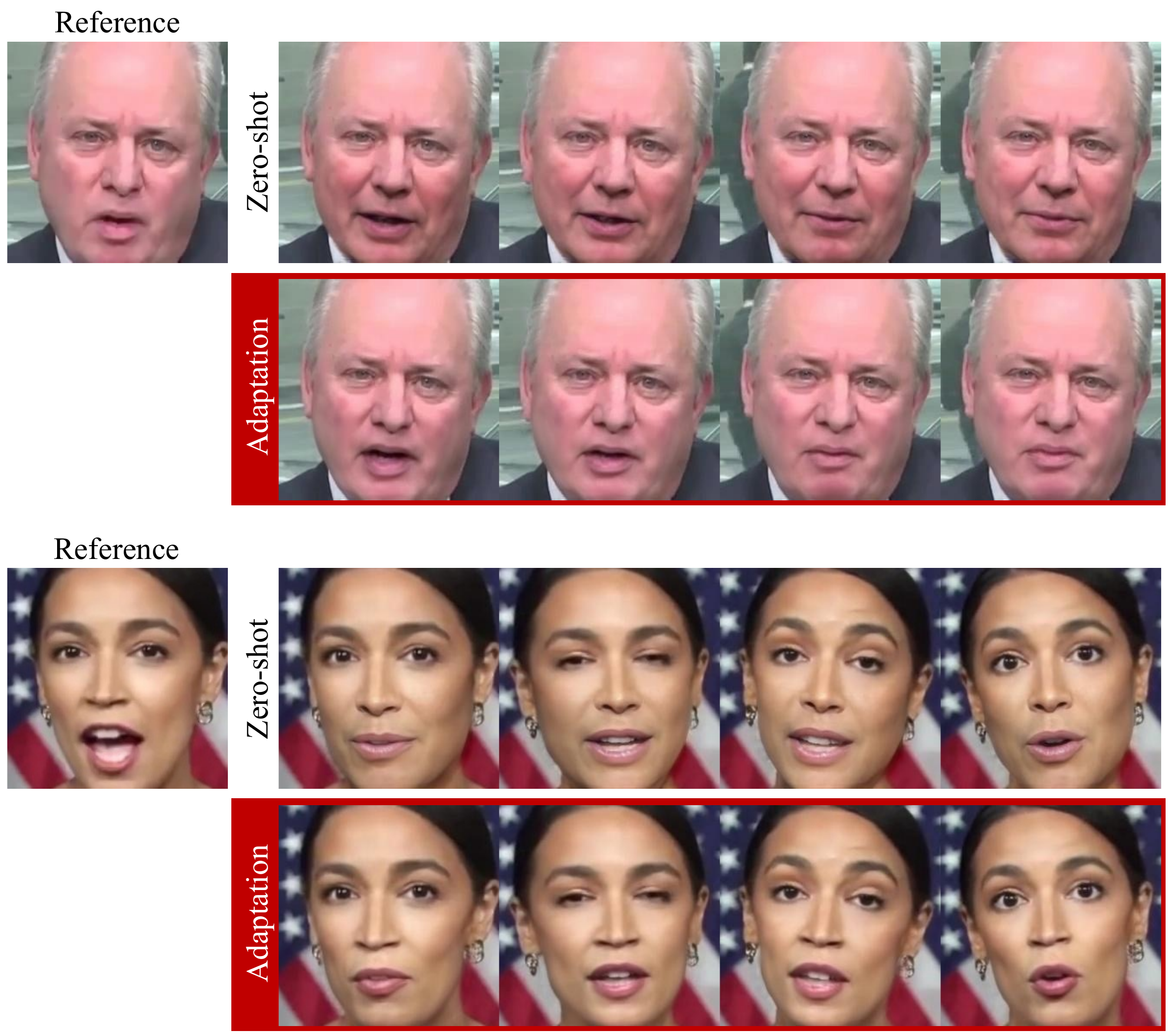}\label{fig:cross_adap}}
    \subfigure[Ablation study on the proposed sync regularizer.]{\includegraphics[width=\linewidth]{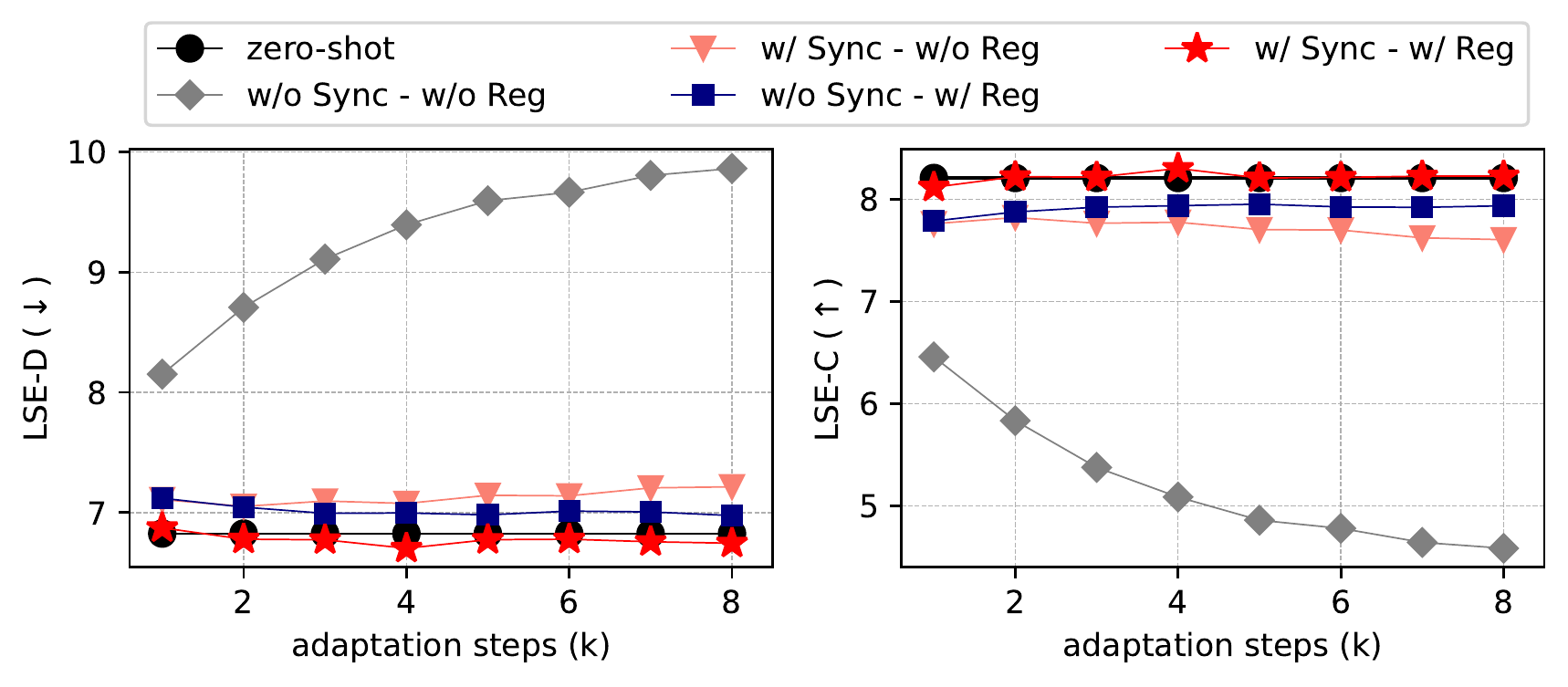}\label{fig:regularizer}}
\end{center}
    \vspace*{-3mm}
    \caption{Experimental results of the proposed adaptation method using the cross-id setting. It improves the person specific visuals while maintaining the lip-sync generality.}
    \label{fig:ablation_unseen}
\end{figure}

\noindent \textbf{Ablation Studies on Unseen Face Adaptation.} \indent We conduct ablation studies on the proposed unseen face adaptation following the same setting in Table \ref{tab:cross}. Additionally, we use 60 seconds of video for each 10 personalized models and 15 audios of 10 seconds from different identities for inference. Figure \ref{fig:regularizer} shows the lip-sync metrics according to the adaptation step. In the cases without the sync regularizer, the models lose the lip-sync generality, in other words, it memorizes the short target video as the adaptation phase proceeds. Introducing the sync regularizer with the sync loss stabilizes the lip-sync metrics and even improves them compared to the zero-shot results. If we use the sync regularizer without the sync loss, lip-sync quality is stabilized, however slightly lower than the zero-shot results due to the lack of ground-truth audio-visual correlation. Audio generalization for unseen face data is maintained even though we used audio from the learning data. Figure \ref{fig:cross_adap} supports the validity of the adaptation method. It shows that visual difference between the zero-shot results and the adaptation results. The zero-shot model can generate an accurate lip motion for the target audio, while it shows a slight difference in person-specific details compared to the reference images. Through the proposed adaptation method, personal-specific lip shape, teeth, and wrinkles are faithfully recovered while maintaining the lip motion of the zero-shot results.

\subsection{User Study}

\begin{table}[tb]
    \caption{Mean Opinion score (MOS) user study results with $95\%$ confidence interval on cross-id setting. The score ranges in 1 to 5. The best score for each metric is in \textbf{bold}.}\label{tab:user}
    \vspace*{-1mm}
    \begin{center}
        \scalebox{0.82}{
            \begin{tabular}{l||c|c|c}
                \toprule
                \multicolumn{1}{c||}{} & \multicolumn{3}{c}{User Study (MOS)} \\
                \multicolumn{1}{l||}{Method} & \multicolumn{1}{c|}{Lip-sync} & \multicolumn{1}{c|}{Face} & \multicolumn{1}{c}{Visual} \\
                \multicolumn{1}{l||}{} & \multicolumn{1}{c|}{Accuracy} & \multicolumn{1}{c|}{Similarity} & \multicolumn{1}{c}{Quality} \\
                \hline
                Wav2Lip \cite{wav2lip} & $3.76 \pm 0.20$ & $2.98 \pm 0.25$ & $2.03 \pm 0.21$ \\
                ATVG \cite{atvg} & $2.19 \pm 0.21$ & $2.45 \pm 0.25$ & $1.54 \pm 0.17$\\
                MakeItTalk \cite{makeittalk} & $2.32 \pm 0.24$ & $3.47 \pm 0.23$ & $2.95 \pm 0.24$  \\
                PC-AVS \cite{pcavs} & $3.28 \pm 0.21$ & $2.55 \pm 0.22$ & $2.51 \pm 0.22$  \\
                \cdashline{1-4}
                \textbf{Ours} & $\textbf{4.01} \pm \textbf{0.16}$ & $3.42 \pm 0.22$ & $3.55 \pm 0.20$\\
                \textbf{Ours (Personalized)} & $3.52 \pm 0.21$ & $\textbf{4.03} \pm \textbf{0.17}$ & $\textbf{3.64} \pm \textbf{0.19}$ \\
                \bottomrule
            \end{tabular}
            }
    \end{center}
    \vspace*{-5mm}
\end{table}
We further conduct a user study based on MOS (Mean opinion score) to compare the perceptual quality of each model, including our zero-shot and personalized model. 5 videos generated by each model in cross-id setting are used for this study. 20 participants scored lip-sync accuracy, face similarity, and visual quality of each video in the range of 1 to 5. As shown in Table \ref{tab:user}, our models outperform all metrics. Specifically, our zero-shot model achieves the highest lip-sync accuracy, and our adaptation model achieves the highest score in face similarity and visual quality with competitive lip-sync accuracy.

\section{Conclusion}
\label{sec:conclusion}
We proposed StyleLipSync, a lip-sync video generative model for arbitrary identity, which leverages expressive lip priors from a pre-trained style-based generator. In contrast to existing lip-sync generative models, we introduce pose-aware masking for lip region by utilizing a 3D parametric mesh predictor, which inherits the pose information in the mask itself. Designing a smooth lip motion by using the moving-average based latent smoothing in the continuous and linear latent space, StyleLipSync can generate temporally consistent lip motion. Furthermore, we propose a few-shot lip-sync adaptation method for a single person who lies in the out-of-distribution, which uses a few seconds of the target person's video. Experimental results show that our StyleLipSync can generate realistic lip-sync video from arbitrary audio even with the zero-shot setting, and the proposed adaptation method enhances the person-specific visual information without losing the lip-sync generality.

\noindent \textbf{Limitation and Future Works.} \indent Since learning audio-visual representation requires a large number of different identities (e.g., Voxceleb2 \cite{vox2}), extending our method to higher resolution ($512 \times 512\uparrow$) is very challenging. We consider it our future work to develop an effective encoding system for a limited number of identities using a style-based generator \cite{stylegan1, stylegan2, stylegan3}. Designing an effective reference encoder to improve lip identity preservation in a zero-shot setting can be another future work.

\noindent \textbf{Ethical Considerations.} \indent Since our method can generate a video of a specific person talking specific words only with a few seconds of video source, it has the potential for misuse. As discussed in \cite{videoretalking}, attaching visual watermarks on the generated videos can be a solution to it.

{\small
\bibliographystyle{ieee_fullname}
\bibliography{main}
}



\end{document}